\documentclass{article}

     \usepackage[preprint]{neurips_2022}

\usepackage[utf8]{inputenc} 
\usepackage[T1]{fontenc}    
\usepackage{hyperref}       
\usepackage{url}            
\usepackage{booktabs}       
\usepackage{amsfonts}       
\usepackage{nicefrac}       
\usepackage{microtype}      
\usepackage{xcolor}  
\usepackage{wrapfig}
\usepackage{graphicx,subfigure}
\usepackage{amsmath}
\usepackage{amssymb}
\usepackage{multirow}
\title{Hierarchical Spherical CNNs with Lifting-based Adaptive Wavelets for Pooling and Unpooling}
\author{%
  Mingxing Xu\thanks{Work was done during Mingxing's visit in LTS4, EPFL, Switzerland, supervised by Prof. Pascal Frossard.} \\
  Shanghai Jiao Tong University\\
  \texttt{xumingxing@sjtu.edu.cn}\\
  \And
  Chenglin Li\\
  Shanghai Jiao Tong University\\
  \texttt{lcl1985@sjtu.edu.cn}\\
  \And
  Wenrui Dai\\
  Shanghai Jiao Tong University\\
  \texttt{daiwenrui@sjtu.edu.cn}\\
  \And
  Siheng Chen\\
  Shanghai Jiao Tong University\\
  \texttt{sihengc@sjtu.edu.cn}
  \And
  Junni Zou\\
  Shanghai Jiao Tong University\\
  \texttt{zoujunni@sjtu.edu.cn}\\
  \And
  Pascal Frossard\\
  École Polytechnique Fédérale de Lausanne (EPFL) \\
  \texttt{pascal.frossard@epfl.ch}\\
   \And
  Hongkai Xiong\\
  Shanghai Jiao Tong University\\
  \texttt{xionghongkai@sjtu.edu.cn}\\
}
\begin{document}
\maketitle
\begin{abstract}\vspace{-0.2cm}
Pooling and unpooling are two essential operations in constructing hierarchical spherical convolutional neural networks (HS-CNNs) for comprehensive feature learning in the spherical domain. Most existing models employ downsampling-based pooling, which will inevitably incur information loss and cannot adapt to different spherical signals (with different spectra) and tasks (dependent on different frequency components). Besides, the preserved information after pooling cannot be well restored by the subsequent unpooling to characterize the desirable features for a task. In this paper, we propose a novel framework of HS-CNNs with a lifting structure to learn adaptive spherical wavelets for pooling and unpooling, dubbed LiftHS-CNN, which ensures a more efficient hierarchical feature learning for both image- and pixel-level tasks. Specifically, adaptive spherical wavelets are learned with a lifting structure that consists of trainable lifting operators (i.e., \textit{update} and \textit{predict} operators). With this learnable lifting structure, we can adaptively partition a signal into two sub-bands containing low- and high-frequency components, respectively, and thus generate a better down-scaled representation for pooling by preserving more information in the low-frequency sub-band. The \textit{update} and \textit{predict} operators are parameterized with graph-based attention to jointly consider the signal's characteristics and the underlying geometries. We further show that particular properties (i.e., \textit{spatial locality} and \textit{vanishing moments}) are promised by the learned wavelets, ensuring the spatial-frequency localization for better exploiting the signal's correlation in both spatial and frequency domains. We then propose an unpooling operation that is invertible to the lifting-based pooling, where an inverse wavelet transform is performed by using the learned lifting operators to restore an up-scaled representation. Extensive empirical evaluations on various spherical domain tasks (e.g., spherical image reconstruction, classification, and semantic segmentation) validate the superiority of the proposed LiftHS-CNN. 
\end{abstract}

\section{Introduction}
Recently, there is an ever-increasing interest in developing spherical CNNs~\cite{boomsma2017spherical,cobb2020efficient,cohen2019gauge,cohen2018spherical,coors2018spherenet,defferrard2020deepsphere,esteves2018learning,esteves2020spin,jiang2019spherical,kondor2018clebsch,mcewen2021scattering,perraudin2019deepsphere}, which generalize classical CNNs to processing ubiquitous spherical data. Examples include omnidirectional images and videos in virtual reality~\cite{coors2018spherenet}, planetary data in climate science~\cite{racah2017extremeweather}, observations of the universe and LIDAR scans~\cite{geiger2013vision,perraudin2019deepsphere}, molecular structures in chemistry~\cite{boomsma2017spherical}. Spherical CNNs can extract features with convolutions defined in a spherical harmonic space~\cite{cobb2020efficient,cohen2018spherical,esteves2018learning,esteves2020spin,kondor2018clebsch} or discrete space~\cite{boomsma2017spherical,cohen2019gauge,defferrard2020deepsphere,perraudin2019deepsphere}, achieving striking performance in many tasks such as classification and semantic segmentation for spherical images. Nevertheless, most of the effort to date has been devoted to designing spherical convolutions that extract features at a single scale, while learning of hierarchical features in the spherical domain is not well explored yet. To address this, hierarchical spherical CNNs (HS-CNNs) are thus demanded for comprehensive feature learning. 

In classical CNNs, pooling and unpooling operations are indispensable for an efficient learning of the hierarchical features, which unfortunately are mostly overlooked in HS-CNNs. Mimicking classical CNNs, and with the aid of the naturally hierarchical structures of some sphere discretization schemes (e.g., icosahedral~\cite{baumgardner1985icosahedral}, equiangular~\cite{driscoll1994computing}, and healpix~\cite{gorski2005healpix} samplings), pooling in HS-CNNs~\cite{defferrard2020deepsphere,jiang2019spherical,perraudin2019deepsphere,shen2021pdo} is typically performed through a downsampling (or mean, max) operation within a local region. However, inspired by the notions of signal processing such as sampling theorem and filtering, simply downsampling (or taking the average or maximum of) a representation will inevitably result in information loss~\cite{saeedan2018detail,williams2018wavelet}, since these operations are rigid without considering the spatial-frequency characteristics of original signals. For example, the sampling theorem suggests that the information can be well preserved after downsampling only if the signals are sufficiently smooth within a limited band. When this is not the case, important information may be lost, which will greatly hamper the information flow across a deep model.
Furthermore, the notion of important information is also varying with signals (with different spectra) or tasks (dependent on different frequency components). Therefore, a desirable pooling should have the capacity to adaptively produce a down-scaled representation that maintains the important information with respect to the data and tasks at hand. On the opposite, existing HS-CNNs widely adopt the padding-based unpooling by trivially inserting zeros for an up-scaled representation~\cite{defferrard2020deepsphere,jiang2019spherical,perraudin2019deepsphere,shen2021pdo}. This, however, cannot well restore the information preserved after pooling, which will inevitably introduce annoying artifacts and discontinuities, and thus degrade the overall performance of HS-CNNs. 

To alleviate the information loss between representations of different scales, we resort to the lifting structure, a widely-adopted technique for multi-resolution analysis in signal processing, to learn adaptive spherical wavelets for constructing advanced pooling and unpooling operations. The spatial-frequency localization of the wavelets enables a better exploitation of the signal's characteristics in both spatial and frequency domains. In particular, the spatial locality is brought by the local compact support, while the smoothness and vanishing moments together contribute to its frequency locality. It is worth mentioning that constructing wavelets on the sphere is a nontrivial task, since the sphere in essence is a non-Euclidean domain~\cite{schroder1995spherical}, let alone the adaptive ones.
The lifting structure~\cite{sweldens1996lifting,sweldens1998lifting} provides us with an efficient way to construct wavelets adaptive to arbitrary domains, while its spatial implementation ensures an easy control of some desirable properties of the resulting wavelets, such as the \textit{spatial locality} and \textit{vanishing moments}. Moreover, an inverse wavelet transform can be easily performed with the backward lifting, where invertibility of the resulting wavelets is theoretically guaranteed. Besides, adaptivity can also be incorporated to adapt wavelets to various data and tasks. 

We are thus motivated to propose a novel framework of HS-CNNs, dubbed LiftHS-CNN, with a lifting structure to learn adaptive spherical wavelets for pooling and unpooling with reduced information loss, which is thus able to ensure a more efficient learning of hierarchical features for both image- and pixel-level tasks in the spherical domain. Specifically, adaptive spherical wavelets are learned with a lifting structure that consists of trainable lifting operators (i.e., \textit{update} and \textit{predict} operators) to adaptively partition a signal into two sub-bands containing low- and high-frequency components. The low-frequency sub-band is encouraged to preserve more information by minimizing the energy of the high-frequency sub-band, which is kept after pooling and helps generate a better down-scaled representation. The \textit{update} and \textit{predict} operators are parameterized with graph-based attention to jointly consider the signal's characteristics and the underlying geometries. We further show that particular properties (i.e., \textit{spatial locality} and primary/dual first-order \textit{vanishing moment} in our case) are promised by the learned wavelets, which ensure the spatial-frequency localization for better exploiting the signal's characteristics. We then propose an unpooling operation that is invertible to the lifting-based pooling, by leveraging an inverse wavelet transform with the learned lifting operators to recover an up-scaled representation. The proposed lifting-based pooling and unpooling are compatible with most existing spherical convolutions to form an HS-CNN, which can be optimized in an end-to-end manner. To evaluate the superiority of the proposed model, we perform extensive experiments on a wide range of spherical  datasets (i.e., SCIFAR10, SMNIST, SModelNet40, and Standford 2D3DS) for spherical image reconstruction, classification, and semantic segmentation. Experimental results validate the effectiveness and efficiency of the proposed LiftHS-CNN. Our contributions can be summarized as follows:
\begin{itemize}
    \item We leverage lifting structure to learn adaptive spherical wavelets for constructing advanced pooling and unpooling operations. Based on them, we further introduce a novel framework of LiftHS-CNNs for a more efficient learning of hierarchical features in the spherical domain.
    \item We implement the lifting operators (i.e., \textit{update} and \textit{predict} operators) with graph-based attention, to jointly consider the characteristics of local signals and the underlying geometries.
    \item We show that desirable properties (i.e., \textit{spatial locality}, and \textit{vanishing moments}) are promised by the learned spherical wavelets, which ensure the spatial-frequency localization.
\end{itemize}

\section{Related Work}\vspace{-0.2cm}

\textbf{Spherical CNNs.} 
Existing models can be categorized into two groups, based on the spaces (harmonic or discrete grid spaces) in which the spherical convolution is defined. \cite{cohen2018spherical,esteves2018learning,esteves2020spin,kondor2018clebsch} define spherical convolutions via spherical harmonic transforms. 
Though preserving excellent rotational equivariance, these models are typically costly and flat, without the capacity to learn hierarchical features. In contrast, \cite{cohen2019gauge,defferrard2020deepsphere,jiang2019spherical,khasanova2017graph,perraudin2019deepsphere,shen2021pdo,yang2020rotation,zhang2019orientation} define convolutions on a discrete grid space (e.g., Healpix and icosahedral grids), striking a trade-off between efficiency and rotational equivariance. For example, \cite{jiang2019spherical, shen2021pdo} adopt parameterized differential operators for feature extraction, while~\cite{defferrard2020deepsphere,khasanova2017graph,perraudin2019deepsphere,yang2020rotation} represent the sphere with a weighted graph and process it with graph CNNs. The natural hierarchy of discrete grids enables hierarchical feature learning, requiring more efficient pooling and unpooling.

\textbf{Hierarchical CNNs.} Pooling and unpooling are ubiquitous in classical CNNs for hierarchical feature learning. Unpooling is typically performed by padding zeros into an up-scaled representation, while various pooling operations are studied. Mean~\cite{lecun1989handwritten} and max pooling~\cite{jarrett2009best} are the most widely-used techniques, which, however, ignore the signal's spatial-frequency characteristics, causing information loss and lack of adaptivity. Mixed pooling~\cite{lee2016generalizing,yu2014mixed} and $L_p$ pooling~\cite{estrach2014signal,gulcehre2014learned} are then proposed to improve them. To preserve more structured details, detail-preserved pooling~\cite{saeedan2018detail} is then introduced. For a more compact representation with less artifacts, wavelet-based pooling operations~\cite{williams2018wavelet,wolter2021adaptive} are proposed. However, the capacity of pooling~\cite{williams2018wavelet} based on fixed Haar wavelets is limited, while the parameterized wavelets~\cite{wolter2021adaptive} are hard to optimize without a theoretical guarantee of some desirable properties, such as \textit{spatial locality}, \textit{vanishing moments} and \textit{invertibility}. The most relevant work to ours is LiftPool~\cite{zhao2021liftpool}, which leverages the invertibility of lifting scheme to improve pooling and unpooling in the 2D Euclidean domain. However, the proposed nonlinear neural network-based lifting operators with the changed down-scaled input inevitably break the \textit{invertibility}, while some desirable properties of the learned transforms are unfortunately not guaranteed and studied. Furthermore, its 2D separable lifting scheme cannot be easily applied to the spherical domain, as directions are not well defined on the sphere. In contrast, we leverage the lifting structure to learn adaptive spherical wavelets whose desirable properties are theoretically guaranteed. The learned operators can also be directly reused for inverse wavelet transform in the unpooling process, thus maintaining the \textit{invertibility}.  

\textbf{Lifting scheme.} The lifting scheme~\cite{sweldens1996lifting,sweldens1998lifting} is a general signal processing technique for constructing  wavelets for both regular and irregular domains. For instance, it is widely used in generating adaptive wavelets for perfect-reconstruction filter banks~\cite{piella2002adaptive}, image coding~\cite{claypoole2003nonlinear}, and denosing~\cite{wu2004adaptive}. Besides, it is also prevalent in constructing wavelets in irregular domains, such as spheres~\cite{schroder1995spherical}, trees~\cite{shen2008optimized}, and graphs~\cite{narang2009lifting}. Recently, there is a surge of interest in integrating the lifting scheme with deep learning~\cite{li2021reversible,rodriguez2020deep,rustamov2013wavelets,xu2022graph} to improve its interpretability and model capacity. To the best of our knowledge, constructing adaptive wavelets on the sphere with lifting has nevertheless not been explored yet, which is non-trivial since both the signal's characteristics and the non-Euclidean geometry should be considered. We resort to graph-based lifting to learn adaptive spherical wavelets, where lifting operators are implemented with graph-based attention to jointly consider both of them.

\section{Lifting-based Hierarchical Spherical CNNs}\vspace{-0.2cm}
In HS-CNNs, hierarchical features are typically extracted by alternating the convolution and pooling operations, while unpooling is further needed for the pixel-level tasks, such as reconstruction. To promote the information flow, it is critical for pooling to produce a down-scaled representation that maintains the most important information, and for unpooling to restore the preserved information to an up-scaled representation that characterizes the desirable features for a task. This goal is arguably difficult to achieve with the standard pooling (e.g., downsampling, mean or max) and unpooling (e.g., padding). In this section, we first describe the classic lifting structure with handcrafted spherical wavelets, and then introduce the proposed graph attention-based lifting for adaptive spherical wavelets. We finally construct the lifting-based pooling and unpooling, as well as the proposed LiftHS-CNN. 
\begin{figure}
  \centering
  \subfigure[Lifting scheme]{
 \begin{minipage}{0.6\linewidth}
  \centering
 \includegraphics[width=\columnwidth]{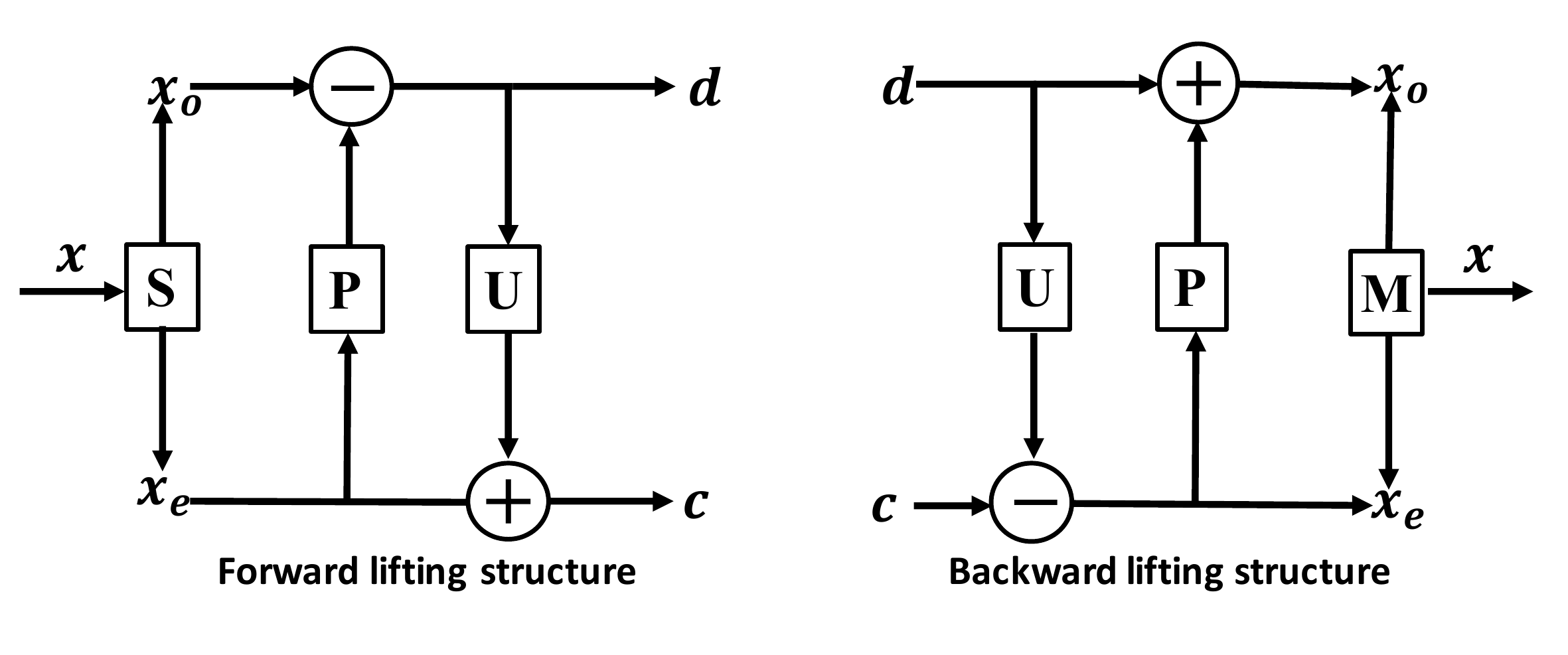}
 \end{minipage}}\subfigure[Lifting on 1D grids and sphere]{
 \begin{minipage}{0.4\linewidth}
  \centering
 \includegraphics[width=\columnwidth]{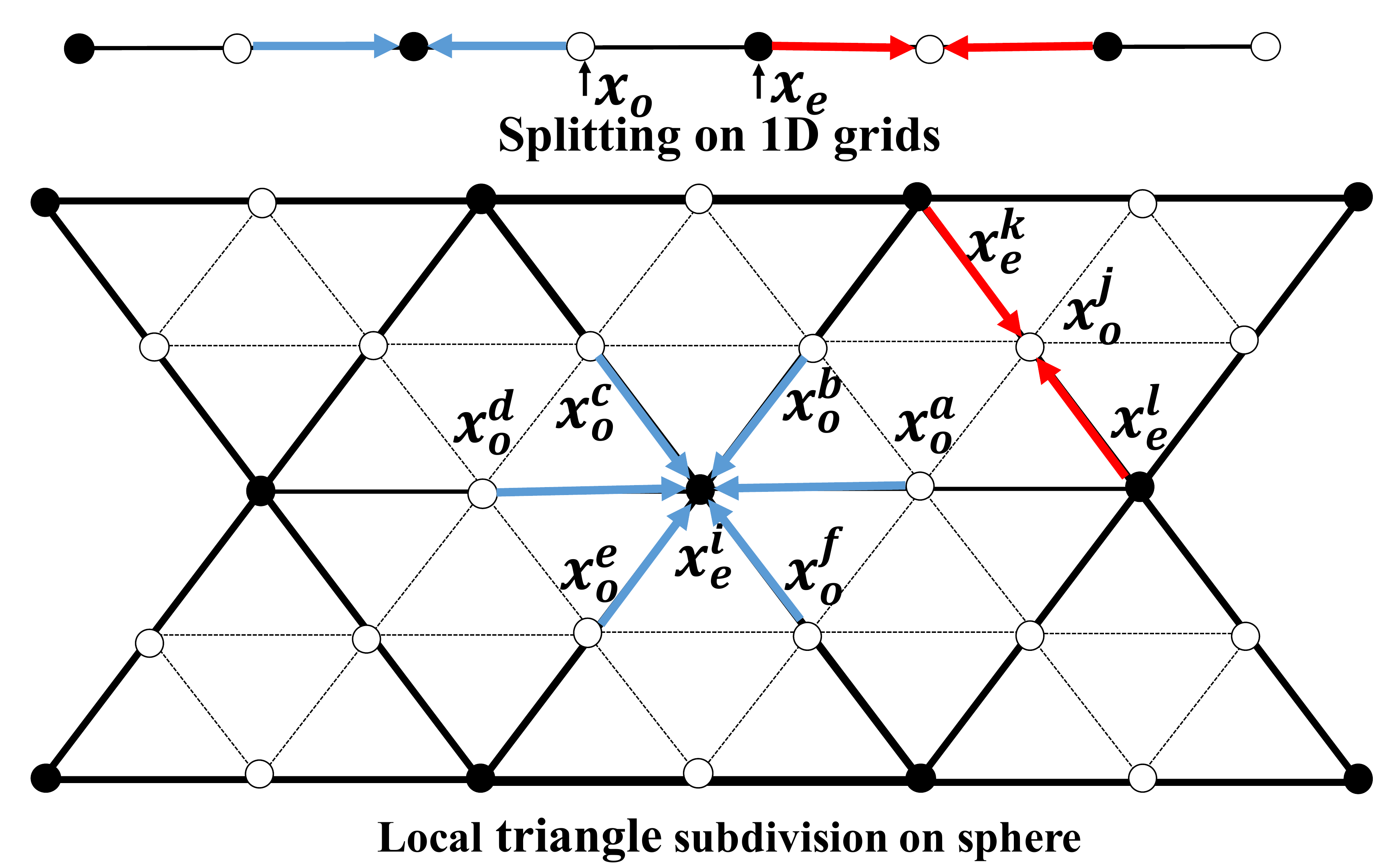}
 \end{minipage}
 }\vspace{-0.25cm}
  \caption{(a) A lifting scheme includes forward and backward lifting structures. The forward lifting structure includes three operations: signal splitting ($\mathbf{S}$), prediction (with \textit{predict} operator $\mathbf{P}$ and subtraction) and update (with \textit{update} operator $\mathbf{U}$ and summation). It is inverted by the backward lifting structure, where $\mathbf{M}$ denotes signal merging. (b) Illustration of lifting for signals lying on 1D grids and sphere with icosahedron-based subdivision, where dots and circles represent even and odd elements, while red and blue arrows denote prediction and update operations, respectively.}
  \label{fig1}\vspace{-0.5cm}
\end{figure}

\subsection{Lifting Scheme for Handcrafted Spherical Wavelets}
In the lifting scheme as shown in Fig.~\ref{fig1}(a), the forward lifting structure consists of three elementary operations, i.e., splitting, prediction and update. Given a single-channel 1D signal $\mathbf{x}\in \mathbb{R}^{N}$ with an even number $N$, as shown in Fig.~\ref{fig1}(b), it is first split into two complementary partitions, i.e., the odd $\mathbf{x}_o\in\mathbb{R}^{N/2}$ and the even $\mathbf{x}_e\in\mathbb{R}^{N/2}$. For each odd component $\mathbf{x}_o^j$ indexed by $j$, we perform the prediction operation based on its neighboring two even components, which subtracts $\mathbf{x}_o^j$ with the estimation to generate the residual $\mathbf{d}^j=\mathbf{x}_o^j-\sum_{j\sim i}\mathbf{P}_{ji}\mathbf{x}_e^i$, where $P_{ji}$ is the \textit{predict} weight. A down-scaled approximation of $\mathbf{x}$ is then obtained as $\mathbf{c}^i=\mathbf{x}_e^i+\sum_{i\sim j}\mathbf{U}_{ij}\mathbf{d}^j$, by updating each even component $\mathbf{x}_e^i$ with the residual $\mathbf{d}^j$, where $U_{ij}$ is the \textit{update} weight. A critically-sampled wavelet transform is implicitly and efficiently performed with this structure, where in frequency domain the signal's sub-band splitting and downsampling are simultaneously realized, i.e., $\mathbf{d}$ and $\mathbf{c}$ carry the high- and low-frequency components, respectively. The resulting wavelet and its properties (e.g., \textit{spatial locality}, \textit{vanishing moments}) are entirely controlled by the \textit{predict} and \textit{update} operators. The inverse wavelet transform can be performed with the backward lifting structure as shown in Fig.~\ref{fig1}(a). It is worth noting that, ideally, the splitting should bipartite $\mathbf{x}$ to make the odd and even partitions highly correlated. By doing so, the correlation between neighboring odd and even components can be further leveraged by the \textit{predict} and \textit{update} operators to attenuate $\mathbf{d}$ while producing a more informative $\mathbf{c}$. 

The lifting scheme has also been applied on the sphere for handcrafted spherical wavelets, based on the geodesic construction of spheres with triangle subdivision~\cite{schroder1995spherical}. To ease the presentation, we take a local triangle subdivision of the sphere in Fig.~\ref{fig1}(b) as an example.
The splitting is naturally done according to the hierarchical subdivision structure (though not ideally bipartite), to partition signals lying on the coarser-level grids as even and those on the remaining grids as odd. The prediction and update operations can then be extended to the sphere. For example, the even components $\mathbf{x}_e^k$ and $\mathbf{x}_e^l$ are used to predict $\mathbf{x}_o^j$, while the residuals of the six odd elements $\mathbf{x}_o^{a\sim f}$ are adopted to update $\mathbf{x}_e^i$, as  
$\mathbf{d}^j=\mathbf{x}_o^j-\mathbf{P}_{jk}\mathbf{x}_e^k-\mathbf{P}_{jl}\mathbf{x}_e^l$, $\mathbf{c}^i=\mathbf{x}_e^i+\sum_{i\sim t, t \in \{a,\cdots, f\}}\mathbf{U}_{it}\mathbf{d}^t$.
Consequently, spherical wavelets with desirable properties can be analytically constructed. For instance, wavelets with the first-order \textit{vanishing moment} can be constructed by setting $\mathbf{P}_{jk}=\mathbf{P}_{jl}=1/2$. This setting has properly considered the fact that the geodesic distance between $\mathbf{x}_o^j$ and $\{\mathbf{x}_e^k,\mathbf{x}_e^l\}$ is equal, such that for constant input signals, the detail (wavelet) coefficients (i.e., $\mathbf{d}$) will always be zeros. To obtain stable wavelets, a dual first-order vanishing moment is preferred~\cite{rustamov2013wavelets}, which requires the update operation to maintain the first-order moment (i.e., mean) of the input signal in its approximation $\mathbf{c}$. We denote the coarsest-level subdivision of the sphere as $\tilde{S}_0$ and the $l$-th level subdivision as $\tilde{S}_l$. If further defining the integral on $\tilde{S}_l$ as $\int_{\tilde{S}_l}\mathbf{x}=\sum_{i\in \tilde{S}_l}\mathbf{x}^i$, the dual first-order vanishing moment then requires $\textit{mean}(\sum_{i\in \tilde{S}_{l}}\mathbf{c}^i)= \textit{mean}(\sum_{i\in \tilde{S}_{l+1}}\mathbf{x}^i)$.

\subsection{Graph Attention-based Lifting for Adaptive Spherical Wavelets}

The above spherical wavelet transform is fixed with a handcrafted design of the \textit{predict} weights, resulting in a rigid frequency sub-band splitting. Though considering geometry of the sphere (i.e., almost invariant everywhere with the triangle subdivision), it cannot adapt to the signal's spatial-frequency characteristics and the frequency preference of a particular task. Thus, we propose in this subsection to learn adaptive spherical wavelets with graph-based attention, which can flexibly split a signal into two sub-bands by jointly considering both the characteristics of signals and tasks. The lifting operators are parameterized with graph attention, and can be optimized in an end-to-end manner. Moreover, specific designs are introduced to theoretically guarantee the \textit{spatial locality} and the first-order \textit{vanishing moment} of the resulting wavelets. These designs together ensure the spatial-frequency localization of the learned spherical wavelets for better exploiting the signal's characteristics in both spatial and frequency domains. Note that, to better restore up-scaled representations, the update-first lifting scheme is usually adopted (e.g., in classical image coding~\cite{claypoole2003nonlinear,li2021reversible,rustamov2013wavelets}), which is slightly different from Fig.~\ref{fig1}(a) in that the order of prediction and update operations is switched. Namely, in the forward lifting, update is performed first and then the prediction, while in the backward lifting, prediction implements first and then the update. With this update-first lifting, more neighbors can be incorporated for the prediction (i.e., determining the wavelets), such that both the \textit{predict} and \textit{update} operators can be utilized in the backward lifting even without transmitting the detail coefficients, i.e., by setting $\mathbf{d}=\mathbf{0}$.    

\begin{wrapfigure}{r}{6.485cm}
  \centering
 \includegraphics[width=6.485cm]{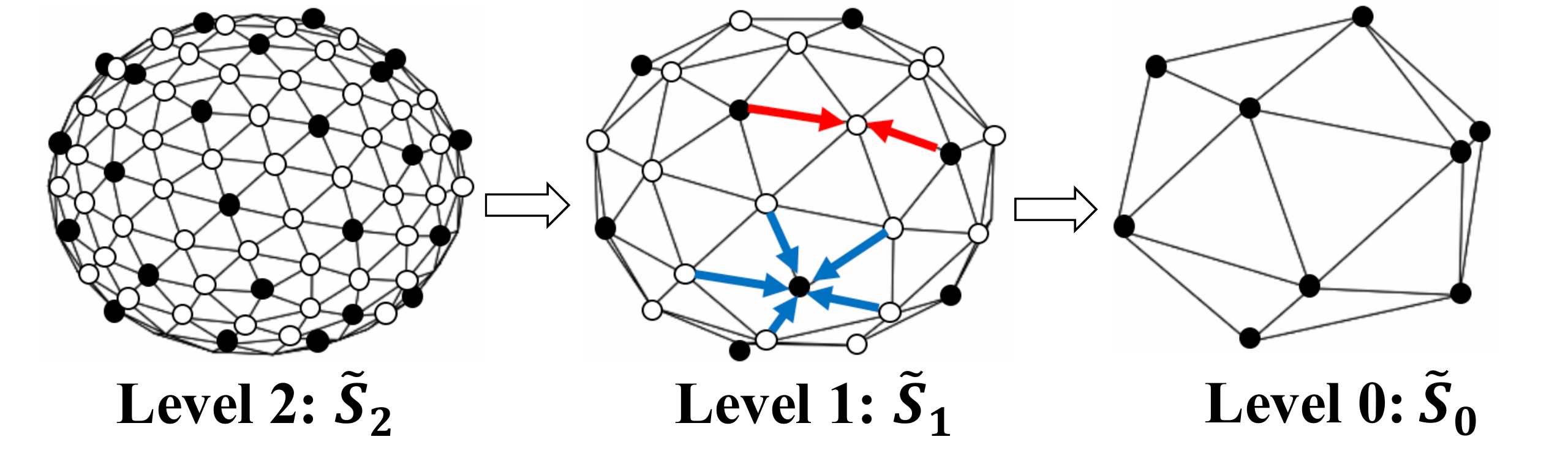}\vspace{-0.25cm}
  \caption{Illustration of first three levels of icosahedron-based subdivision on sphere. At each level, grids are partitioned into even (dots) and odd (circles) partitions. When pooling from finer- to coarser-levels (e.g., $\tilde{S}_1 \to \tilde{S}_0$), signals lying on the dots are updated (indicated by blue arrows) and preserved, while those on the circles are predicted (indicated by red arrows) with the resulting residuals dropped. The \textit{predict} and \textit{update} weights are reused in backward lifting to restore finer representations.}
 \label{fig2}\vspace{-0.5cm}
\end{wrapfigure}
\textbf{Graph construction with subdivisions of sphere.} We adopt the icosahedron-based subdivision of sphere for lifting, as shown in Fig.~\ref{fig2}, since it is the most uniform and accurate discretization of a sphere~\cite{baumgardner1985icosahedral,jiang2019spherical}, which preserves geometry of the sphere. Besides, the geodesic distance between any two discretized nodes is almost invariant everywhere, which eases learning of the lifting operators and enables weight sharing. Starting with the unit icosahedron, each face is recursively divided into four smaller triangles by putting three additional points at the mid-point of each edge. The newly generated points are then projected onto the unit sphere. By repeating this process, we obtain a sequence of subdivisions of the sphere, where the number of nodes of a level-$l$ subdivision is $|\tilde{S}_l|=10\times4^l+2$. This subdivision process provides a natural splitting scheme of even and odd partitions for lifting. Specifically, for $\tilde{S}_{l}$, the nodes that belong to its precedent subdivision $\tilde{S}_{l-1}$ constitute the even partition, while the remaining nodes form the odd partition. At each level $l$, a graph can be accordingly constructed to represent the geometric structure of sphere, and specify connectivity between nodes from the even and odd partitions. In Fig.~\ref{fig2}, we adopt the inherent graph of each triangle subdivision due to its simplicity, which has an approximate bi-partition with the aforementioned splitting. 
This graph construction incorporates a limited number of neighbors in the \textit{predict} and \textit{update} operators, which provides good spatial locality, but limits the possible orders of vanishing moments of the learned wavelets. More advanced graph construction and splitting scheme will be left for future study.

\textbf{Graph attention-based lifting.} To further incorporate the signal's characteristics, graph attention~\cite{velivckovic2017graph} is adopted to learn the \textit{predict} and \textit{update} operators for constructing adaptive spherical wavelets. These learned lifting operators can be directly used in the backward lifting structure to perform the inverse wavelet transform. 
As will be discussed in detail, by restricting the attention on local sub-graphs, the \textit{spatial locality} of resulting wavelets is guaranteed, while the primary first-order \textit{vanishing moment} is further ensured by constraining the learned lifting operators.

We denote $\mathbf{A}_0, \mathbf{A}_1,\cdots, \mathbf{A}_l$ as the adjacency matrices of the graphs constructed for $\tilde{S}_0,\tilde{S}_1,\cdots, \tilde{S}_l$, each of which consists of ones and zeros describing connectivity of the graph. For example, the nonzero elements in the $i$-th row of $\mathbf{A}_l$ indicate which nodes are connected with the $i$-th node in a level-$l$ graph $\tilde{S}_l$. We further denote the features (or representations) on $\tilde{S}_{l}$ as $\mathbf{X}^l\in\mathbb{R}^{|\tilde{S}_l|\times f^l}$. After splitting, features of the even partition can be grouped as $\mathbf{X}_e^l\in{\mathbb{R}^{|\tilde{S}_{l-1}|\times f^l}}$, while features of the remaining odd partition collected as $\mathbf{X}_o^l\in{\mathbb{R}^{n_l\times f^l}}$. Here, $|\tilde{S}_l|$ denotes the number of nodes, $f^l$ is the feature dimensionality, and $n_l = |\tilde{S}_{l}|-|\tilde{S}_{l-1}|$ for simplicity. Hence, $\mathbf{A}_l$ can be reorganized as
\begin{equation}\label{eq3}
   \mathbf{A}_l= 
   \begin{bmatrix}
    \mathbf{E}_l& \mathbf{M}_l\\
    \mathbf{N}_l& \mathbf{O}_l
   \end{bmatrix},
\end{equation}
where $\mathbf{E}_l\in \mathbb{R}^{|\tilde{S}_{l-1}|\times |\tilde{S}_{l-1}|}$ and  $\mathbf{O}_l\in \mathbb{R}^{ {n_l}\times n_l}$ indicate the sub-graphs that connect nodes within the even and odd partitions, respectively, and $\mathbf{M}_l \in \mathbb{R}^{|\tilde{S}_{l-1}|\times n_l}$ and $\mathbf{N}_l \in \mathbb{R}^{n
_l\times |\tilde{S}_{l-1}|}$ represent the sub-graphs connecting nodes between the even and odd partitions, respectively. Here, $\mathbf{M}_l=\mathbf{N}_l^T$ since $\mathbf{A}_l$ is symmetric. 
The update-first lifting on $\tilde{S}_{l}$ can be written in matrix form as $\mathbf{C}^l=\mathbf{X}_e^l+\mathbf{U}_l\cdot\mathbf{X}_o^l$, and $\mathbf{D}^l=\mathbf{X}_o^l-\mathbf{P}_l\cdot\mathbf{C}^l$, where $\mathbf{C}^l\in{\mathbb{R}^{|\tilde{S}_{l-1}|\times f^l}}$ and $\mathbf{D}^l \in \mathbb{R}^{ {n_l}\times f^l}$ are the approximation and detail coefficient matrices that carry the information of low- and high-frequency sub-bands, respectively, and matrices $\mathbf{U}_l\in \mathbb{R}^{|\tilde{S}_{l-1}|\times n_l}$ and $\mathbf{P}_l\in \mathbb{R}^{n_l\times |\tilde{S}_{l-1}|}$ are the learnable \textit{update} and \textit{predict} operators.

For two nodes $i$ and $j$ on a level-$l$ graph $\tilde{S}_l$, we denote their features as $\mathbf{x}^{i}, \mathbf{x}^{j}\in\mathbb{R}^{1\times f^l}$. With graph attention, the attention weight $t_{ij}$ that attends node $i$ to node $j$ is calculated as
\begin{equation}\label{eq2}
t_{ij}=\sigma([(\mathbf{x}^{i}\mathbf{W}^l_0)\|(\mathbf{x}^{j}\mathbf{W}^l_0)]\cdot[\mathbf{w}^T_1\|\mathbf{w}^T_2]^T)=\sigma(\mathbf{x}^i\mathbf{W}^l_0\mathbf{w}_1+\mathbf{x}^j\mathbf{W}^l_0\mathbf{w}_2),
\end{equation}
where $\|$ denotes concatenation, $\mathbf{W}_{0}^l\in \mathbb{R}^{ f^l \times f^l_0}$ is a trainable weight matrix projecting input features to a hidden space for correlation exploitation, $[\mathbf{w}_{1}^T \| \mathbf{w}^T_{2}]^T\in\mathbb{R}^{2f_0}$ contains weights of a neural network that calculates attention scores with concatenated $[(\mathbf{x}^i\mathbf{W}_0^l)\|(\mathbf{x}^j\mathbf{W}_0^l)]$, and $\sigma$ is an activation function. In practice, graph attention is performed efficiently with sparse matrix operations. Attention on the whole graph can be rewritten in a more compact matrix form, by masking the global attention matrix with 0-1 matrices $\mathbf{M}_l$ and $\mathbf{N}_l$ in Eq.~(\ref{eq3}). The \textit{predict} and \textit{update} operators are thus given by
\begin{equation}\label{eq5}
    \mathbf{U}_l=\mathbf{M}_l\odot\sigma((\mathbf{X}_e^l\mathbf{W}^l_{u0}\mathbf{w}^l_{u1})\oplus(\mathbf{X}_o^l\mathbf{W}^l_{u0}\mathbf{w}^l_{u2})^T), \mathbf{P}_l=\mathbf{N}_l\odot\sigma((\mathbf{X}_o^l\mathbf{W}^l_{p0}\mathbf{w}^l_{p1})\oplus(\mathbf{X}_e^l\mathbf{W}^l_{p0}\mathbf{w}^l_{p2})^T),
\end{equation}
where $\mathbf{W}^l_{u0}, \mathbf{W}^l_{p0}\in \mathbb{R}^{f^l\times f^l_0}$ are learnable projection matrices, $\mathbf{w}^l_{u1}, \mathbf{w}^l_{u2}, \mathbf{w}^l_{p1}, \mathbf{w}^l_{p2} \in \mathbb{R}^{f^l_0}$ are learnable vectors, $\oplus$ is the modified Kronecker product that replaces multiplication of two elements with summation, $\odot$ is the element-wise product between two matrices to perform local attention (\textit{spatial locality}). Note that the learnable weights for \textit{predict} and \textit{update} operators can be further shared to reduce parameters. Post-processing is further performed on the learned \textit{predict} and \textit{update} operators, to guarantee that at least the first-order \textit{vanishing moment} holds by the learned \vspace{-0.1cm} wavelets~\cite{xu2022graph} 
\begin{equation}\label{eq6}
    \mathbf{\hat{U}}_l=\textit{softmax}(\mathbf{{U}}_l), \quad \mathbf{\hat{P}}_l=\textit{softmax}(\mathbf{{P}}_l)/2,
\end{equation}
where $\textit{softmax}(\cdot)$ is performed row-wise. We detail other desired properties of wavelets in Section \ref{sec:property}.

\begin{figure}
  \centering
 \includegraphics[width=1.025\columnwidth]{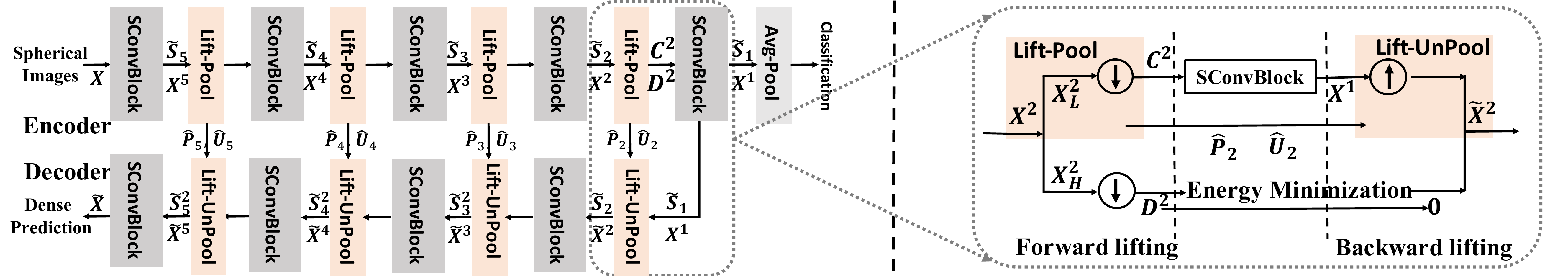}\vspace{-0.175cm}
  \caption{Proposed LiftHS-CNN with lifting-based pooling and unpooling: \textit{i)} Left figure shows the framework for image-level (upper branch only) and pixel-level tasks, where SConvBlock denotes spherical convolutional block for feature extraction and can be chosen w.r.t. tasks. \textit{ii)} Right figure shows an example of lifting-based pooling and unpooling between level-2 and level-1 graphs, where the forward and backward lifting are simplified as their equivalent process in frequency domain, i.e., sub-band splitting and downsampling, and upsampling and frequency sub-band merging, respectively. $\mathbf{X}^2_L$ and $\mathbf{X}^2_H$ are the low- and high-frequency sub-band signals with the same resolution of $\mathbf{X}$.}\vspace{-0.35cm}
  \label{fig3}
\end{figure}

\subsection{Lifting-based  Hierarchical Spherical CNNs}

With the learned $\hat{\mathbf{U}}_l$ and $\hat{\mathbf{P}}_l$, pooling can be performed by implicitly performing wavelet transforms with the graph lifting, and by preserving the approximation coefficients that carry the information of low-frequency sub-band. To obtain $\hat{\mathbf{X}}^{l-1}=\text{LiftPool}(\hat{\mathbf{X}}^{l})$, e.g., the entire process of pooling is 
\begin{equation}\label{eq8}
    \mathbf{X}^l=[\mathbf{X}_e^l,\mathbf{X}_o^l],\quad
    \mathbf{C}^l=\mathbf{X}_e^l+\mathbf{\hat U}_l\cdot\mathbf{X}_o^l, \quad \mathbf{D}^l=\mathbf{X}_o^l-\mathbf{\hat P}_l\cdot\mathbf{C}^l,
\end{equation}
where $\mathbf{C}^l$ is preserved as the down-scaled representation while $\mathbf{D}^l$ is dropped. The energy of $\mathbf{D}^l$ is thus penalized as a regularization term in the final loss function to maximize the information preserved in $\mathbf{C}^l$ ((i.e., by pushing $\|\mathbf{D}^l\|_2\rightarrow 0$)). Besides, to make the learned wavelets have dual first-order \textit{vanishing moment} with improved stability, the first-order moment (i.e., mean) of $\mathbf{X}^l$ is forced to be retained with another regularization term (i.e., $\|\text{Mean}(\mathbf{X}^l)-\text{Mean}(\mathbf{C}^l)\|_2$). 

For dense prediction tasks with the encoder-decoder architecture in Fig.~\ref{fig3}, unpooling is required to restore up-scaled representations, i.e., $\hat{\mathbf{X}}^l=\text{LiftUnPool}(\hat{\mathbf{X}}^{l-1})$. Here, we leverage the inverse wavelet transform for up-scaling, maintaining the perfect information recovery with respect to the preserved low-frequency sub-band after pooling (we detail this property in Section \ref{sec:property}). Specifically, the learned \textit{predict} and \textit{update} operators $\mathbf{\hat P}_l$ and $\mathbf{\hat U}_l$ are reused in the backward lifting structure to perform the inverse wavelet transform for restoring $\mathbf{\hat{X}}^l$ from $\mathbf{\hat{X}}^{l-1}$, leading to the unpooling
\begin{equation}
\mathbf{\hat{X}}^{l}_o=\mathbf{\hat{P}}_l\cdot\mathbf{\hat{X}}^{l-1}, \quad \mathbf{\hat{X}}^{l}_e=\mathbf{\hat{X}}^{l-1}-\mathbf{\hat{U}}_l\cdot\mathbf{\hat{X}}^{l}_o,
\end{equation}
where $\mathbf{\hat{X}}^{l}_o$ and $\mathbf{\hat{X}}^{l}_e$ are merged to form the up-scaled representation $\mathbf{\hat{X}}^{l}$ for further processing.
To train the model, the final loss function is: $\text{loss}=\text{loss}_\text{task}+\lambda\|\mathbf{D}^l\|_2+\gamma\|\text{Mean}(\mathbf{X}^l)-\text{Mean}(\mathbf{C}^l)\|_2$, where the $\text{loss}_\text{task}$ is the task-related loss (e.g., cross entropy loss for classification, and mean square error loss for reconstruction), and $\lambda$ and $\gamma$ are the hyper-parameters for the two regularization terms. \vspace{-0.15cm}

\section{Properties and Complexity}\label{sec:property}\vspace{-0.2cm}
We discuss here some desirable properties of the adaptive spherical wavelets, the information recovery capability and computation complexity of the proposed lifting-based pooling and unpooling.

\textbf{Spatial locality.} The \textit{spatial locality} of resulting wavelets is a direct result of our design, where update and prediction operations are performed with one-hop neighbors on the graph. Therefore, the wavelets realized with the lifting structure (i.e., Eq.~(\ref{eq8})) consisting of a single update and prediction operations are localized within two-hops on the graph~\cite{xu2022graph}. More lifting steps (i.e., prediction and update) in a single forward lifting structure can be iterated down to progressively enlarge the \textit{spatial locality} and possibly improve \textit{vanishing moments} for more advanced spherical wavelets.

\textbf{Primary/dual first-order vanishing moments.} The primary first-order \textit{vanishing moment} requires that for constant input signals, information can be perfectly preserved in the approximation coefficients, i.e., the detail coefficients are all zeros. With the proposed lifting scheme in Eq.~(\ref{eq8}), the resulting wavelets guarantee the primary first-order \textit{vanishing moment}. Please see Appendix B.1 for the proof. The dual first-order \textit{vanishing moment} is enforced by the last regularization term in the loss function, which constrains the first-order moment of input signals after the wavelet transforms. 

\textbf{Perfect recovery of preserved information after pooling.} To better understand the information flow across the entire forward and backward lifting scheme, we interpret the lifting process in frequency domain, with low- and high-pass filtering, and downsampling and upsampling operations~\cite{mallat1999wavelet} as shown in the right half of Fig.~\ref{fig3}. The forward lifting process realizes a critically-sampled wavelet transform by performing filtering and downsampling simultaneously, which can be disentangled into two processes. \textit{i)} The input signal $\mathbf{X}^2$ on level-2 graph $\tilde{S}_{2}$ is firstly filtered by low- and high-pass filters, and partitioned into a low-frequency sub-band $\mathbf{X}^2_L$ and a high-frequency sub-band $\mathbf{X}^2_H$ with the same resolution of $\mathbf{X}^2$. \textit{ii)} Downsampling is then performed for both sub-band signals to obtain the down-scaled approximation $\mathbf{C}^2$ and detail $\mathbf{D}^2$. Similarly, the backward lifting process that realizes an inverse wavelet transform can be disentangled as upsampling and sub-band signal merging processes. It can be proved that the information of $\mathbf{X}_L^2$ is perfectly restored in $\mathbf{\tilde{X}}^2$ through the proposed lifting-based unpooling with the inverse wavelet transform in backward lifting, except for being processed by the spherical convolutional layers in between (i.e., SConvBlock). In contrast, such information cannot be restored with the downsampling-based pooling and padding-based unpooling due to frequency aliasing and artifacts. Please see Appendix B.2 for the introduction of frequency domain  interpretation of the entire lifting process, and the proof for the perfect recovery property.

\textbf{Computation/parameter complexity.}
The lifting-based pooling may slightly bring additional computation and parameter complexity. To get $\mathbf{X}^{l-1}\in\mathbb{R}^{|\hat{S}_{l-1}|\times f^l}$ from $\mathbf{X}^l\in\mathbb{R}^{|\hat{S}_{l}|\times f^l}$ with pooling, graph attention in Eqs.~(\ref{eq5}) and (\ref{eq6}) is performed along the edges of $\mathbf{M}_l=\mathbf{N}^T_l$ to obtain $\mathbf{\hat{U}}_l$ and $\mathbf{\hat{P}}_l$. We denote the number of nonzero elements in $\mathbf{M}_l$ as $|\mathbf{M}_l|$, where $|\mathbf{M}_l|\ll |\hat{S}_{l-1}|\times |\hat{S}_{l}|$ due to sparsity of the constructed graphs. The total computation complexity of the lifting structure with a single update and prediction operation in Eq.~(\ref{eq8}) is then $\mathbf{O}(|\hat{S}_{l}|f^lf^l_0+|\mathbf{M}_l|f^l)$, where $f^l_0$ is the dimensionality of hidden space in Eq.~(\ref{eq2}). This computation complexity is linearly with the number of nodes and edges in the graph, and thus scalable to high-resolution spherical signals. The additional parameter complexity is $\mathbf{O}(f^lf^l_0)$, so the proposed learnable lifting structure is also lightweight. \vspace{-0.15cm}

\section{Experiments}\vspace{-0.2cm}
We evaluate the proposed LiftHS-CNNs on various tasks on benchmark spherical datasets. Specifically, spherical image reconstruction on SCIFAR10 aims to directly verify the information preservation ability of LiftHS-CNN with lifting-based adaptive wavelet for pooling and unpooling, which can be directly reflected by reconstruction error. We then evaluate the model in spherical image classification on SMNIST and SModelNet40 datasets, to show that more task-related information can be preserved by the proposed lifting-based pooling for better discriminality. We further apply the model to spherical image segmentation on Standford 2D3DS dataset, to demonstrate the superiority in dense pixel-level prediction tasks. We end up with the ablation study to better explore the model. 
We briefly describe here for each experimental setting. Please see Appendix A for detailed description. 

\begin{table}[!]
    \centering
    \setlength\tabcolsep{3pt}
     \caption{Comparison of different models for spherical image reconstruction on SCIFAR10.}\vspace{-0.2cm}
    \resizebox{1.0\columnwidth}{!}{\begin{tabular}{c|cccccc|cccccc}
        \hline\hline
          &\multicolumn{6}{|c|}{Training}&\multicolumn{6}{|c}{Test}\\
          \hline
          Models&DownSamp&Mean&Max&LiftHS-CNN$^*$&LiftHS-CNN$_H$&LiftHS-CNN&DownSamp&Mean&Max&LiftHS-CNN$^*$&LiftHS-CNN$_H$&LiftHS-CNN\\
         MSE&0.0031&0.0038&0.0038&0.0028&0.0023& \textbf{0.0019} &0.0030&0.0036&0.0039&0.0025&0.0021& \textbf{0.0017}\\
         \hline\hline
    \end{tabular}}\vspace{-0.5cm}
    \label{tab1}
\end{table}

\begin{figure}
  \centering
\includegraphics[width=1.0\columnwidth]{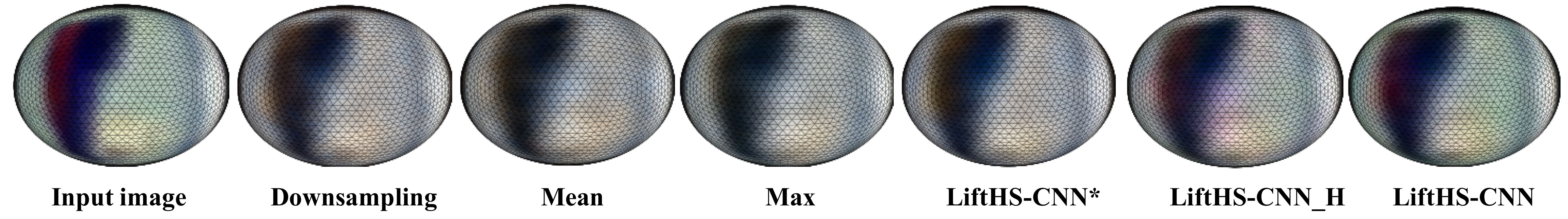}\vspace{-0.425cm}
  \caption{Visualization of spherical images reconstructed by different models on test dataset.}\vspace{-0.6cm}
  \label{fig4}
\end{figure}

\textbf{5.1 Reconstruction of Spherical Images}

\textbf{Experimental setting.} 
Following the common practice~\cite{cohen2018spherical,jiang2019spherical}, we construct SCIFAR10 by projecting images in CIFAR10~\cite{krizhevsky2009learning} onto a unit sphere. SCIFAR10 consists of 50000 training and 10000 test color images, which are more realistic than those in MNIST and thus more challenging to reconstruct. Input signals are sampled with level-4 icosahedron-based subdivision grids, and with 3 channels (i.e., RGB) each of which is normalized into $[0,1]$. 
The encoder-decoder architecture with 4 pooling and unpooling operations is adopted (similar to the architecture in Fig.~\ref{fig3}, but with the maximum level of graph being 4).  For feature extraction at each level (i.e, SConvBlock in Fig.~\ref{fig3}), we employ the same feature extractor built upon MeshConv as~\cite{jiang2019spherical}, due to its effectiveness and high efficiency. Hereafter, this feature extractor is adopted for feature extraction without further specification. 
To validate the superior information preservation ability of the proposed pooling and unpooling, we adopt the same backbone with different pooling (i.e., downsampling, mean, and max) and unpooling (i.e., padding). LiftHS-CNN is the proposed model, LiftHS-CNN$_H$ denotes the same model with handcrafted spherical wavelets, and LiftHS-CNN$^*$ denotes the model with the proposed pooling and padding-based unpooling. Mean square error (MSE) is used to measure reconstruction performance. 

\textbf{Results.}
The MSE loss on training and test datasets are reported in Table~\ref{tab1}. Compared to standard pooling, the converged training MSE loss is reduced when incorporating the proposed lifting-based pooling (i.e., liftHS-CNN$^*$). Furthermore, the performance is largely improved with pooling and unpooling based on both handcrafted and the proposed adaptive wavelets (i.e., LiftHS-CNN$_H$ and LiftHS-CNN). These results imply that model capacity for information preservation is consistently improved by the proposed pooling and unpooling with adaptive spherical wavelets. The generalization ability is further shown by the reduced MSE loss on test dataset, which is consistent with the results on training set, showing that the proposed lifting structures can learn meaningful spherical wavelets that can be generalized to unseen images by preserving more information. We further show visualization results of different models in Fig.~\ref{fig4}, and more visualization results are given in Appendix D.

\textbf{5.2 {Classification of Spherical Images}}

\textbf{Experimental setting.} Two datasets, SMNIST and SModelNet40, are employed to evaluate the proposed model. SMNIST is constructed from MNIST~\cite{lecun1989handwritten} in the same way as SCIFAR10. MNIST consists of 60000 training and 10000 test images presenting handwritten digits. Since SMNIST is relatively easy, we further construct SModelNet40 dataset by transforming geometries of the 3D CAD models in ModelNet40~\cite{wu20153d} to spherical signals. The architecture shown in the upper-left of Fig.~\ref{fig3} is adopted for spherical signal classification. Specifically, for SMNIST, the model with 2 pooling operations from the maximum level of 4 to minimum level of 2 is adopted, while the model with 3 pooling operations from the maximum level of 5 to minimum level of 2 is leveraged for SModelNet40 as it is more challenging. We compare with prevalent spherical CNNs, including S2CNN~\cite{cohen2018spherical}, SphCNN~\cite{esteves2018learning}, FFS2CNN~\cite{kondor2018clebsch}, GaugeCNN~\cite{cohen2019gauge}, SWSCNN~\cite{esteves2020spin}, EGSCNN~\cite{cobb2020efficient}, e$S^2$CNN~\cite{shen2021pdo}, Deepsphere~\cite{defferrard2020deepsphere}, and UGSCNN~\cite{jiang2019spherical}. Note that e$S^2$CNN, Deepsphere, and UGSCNN are hierarchical models, and we adopt the same architecture as UGSCNN except for a lifting-based pooling. 

\begin{table}
    \setlength\tabcolsep{2pt}
    \centering
    \caption{Classification accuracy of different models on SMNIST.}\vspace{-0.15cm}
    \resizebox{1.0\columnwidth}{!}{
    \begin{tabular}{c|ccccccccc}
    \hline\hline
         Models& S2CNN~\cite{cohen2018spherical}& FFS2CNN~\cite{kondor2018clebsch}& GaugeCNN~\cite{cohen2019gauge}& SWSCNN~\cite{esteves2020spin}& EGSCNN~\cite{cobb2020efficient}& 
         e$S^2$CNN~\cite{shen2021pdo}& Deepsphere~\cite{defferrard2020deepsphere}& UGSCNN~\cite{jiang2019spherical}&LiftHS-CNN \\
         Acc (\%)& 95.59& 96.4& 99.43& 99.37& 99.35& 99.44& 98.4& 99.23& \bf{99.60} \\
         $\#$Params& 58k& 286k& 182k& 58k& 58k& 73k& 75k& 62k& 66k \\
    \hline\hline
    \end{tabular}}\vspace{0.05cm}
    \label{SMNIST}
    \setlength\tabcolsep{4pt}
    \centering
    \caption{Classification accuracy of different models on SModelNet40.}\vspace{-0.15cm}
    \resizebox{0.7\columnwidth}{!}{
    \begin{tabular}{c|ccccc}
    \hline\hline
         Models& SphCNN~\cite{esteves2018learning}& SWSCNN~\cite{esteves2020spin}& Deepsphere~\cite{defferrard2020deepsphere}& UGSCNN~\cite{jiang2019spherical}& LiftHS-CNN\\
         Acc (\%)& 89.3& 90.1& 89.47& 90.5& \bf{91.37}\\
    \hline\hline
    \end{tabular}}\vspace{-0.25cm}
    \label{SModelNet40}
\end{table}
\begin{table}[]
    \centering
    \setlength\tabcolsep{3pt}
     \caption{Performance of different models for spherical image segmentation on Standford 2D3DS.}\vspace{-0.15cm}
    \resizebox{0.9\columnwidth}{!}{\begin{tabular}{c|ccccccccc}
        \hline\hline
         Models& Unet~\cite{ronneberger2015u}& GaugeCNN~\cite{cohen2019gauge}& HexRUNet~\cite{zhang2019orientation}&SWSCNN~\cite{esteves2020spin}]&e$S^2$CNN~\cite{shen2021pdo}&UGSCNN~\cite{jiang2019spherical}&LiftHS-CNN\\
         mIou& 0.3587&0.3940&0.433&0.419&0.446&0.3829&\bf{0.4534} \\
         mAcc (\%)& 50.8& 55.9& 58.4& 55.6& 60.4& 54.65& \bf{61.87}\\
         \hline\hline
    \end{tabular}}\vspace{-0.15cm}
    \label{2d3ds}
\end{table}
\textbf{Results.} Different from reconstruction tasks, image classification requires models to preserve  information with a higher discriminality. Classification results on the two datasets are shown in Tables~\ref{SMNIST} and ~\ref{SModelNet40}. We achieve state-of-the-art performance on both datasets, compared to other baselines. Notably, we outperform UGSCNN by 0.37\% on SMNIST and 0.87\% on SModelNet40. These results verify the superiority of the proposed pooling in adaptively preserving important information.


\textbf{5.3 Semantic Segmentation of Spherical Images}

\textbf{Experimental setting.}
We adopt the challenging Stardford 2D3DS dataset~\cite{armeni2017joint} containing 1413 RGB-D equirectangular images with 13 semantic classes that are collected in 6 different areas. It is officially split into 3 folds for cross validation. We compare with prevalent models, including Unet~\cite{ronneberger2015u}, GaugeCNN~\cite{cohen2019gauge},  HexRUNet~\cite{zhang2019orientation}, SWSCNN~\cite{esteves2020spin}], e$S^2$CNN~\cite{shen2021pdo}, UGSCNN~\cite{jiang2019spherical}. 

\textbf{Results.} The performance are shown in Table \ref{2d3ds}. We outperform all the models, and achieve state-of-the-art performance on this challenging dataset. Notably, compared with UGSCNN that employs a similar architecture but with downsampling-based pooling and padding-based unpooling, we achieve significant improvement on both metrics (e.g., 0.0705 and 7.22\% on mIou and mAcc, respectively), indicating the effectiveness of adaptively preserving important information and restoring finer-level features by leveraging the correlation structures encoded with learned spherical wavelets.

\textbf{5.4 Ablation Study}

We further perform ablation study to verify the superiority of proposed LiftPool and LiftUnPool, compared with prevalent pooling (i.e., downsampling, mean and max) and unpooling (i.e., padding) operations, and handcrafted spherical wavelets constructed by fixed \textit{predict} and \textit{update} operators (i.e., LiftHS-CNN$_H$). We adopt for each task the same backbone as UGSCNN, except for different pooling and unpooling operations. The superiority of LiftPool can be validated by results shown in Table~\ref{tab5}, where we outperform all the standard pooling and the one with handcrafted spherical wavelets. Mean and max poolings are better than downsampling for classification tasks, probably because more information loss and aliasing are induced by downsampling. This implies the necessity of advanced pooling for HS-CNNs. The effectiveness of our framework with LiftPool and LiftUnPool is also clearly shown by results in Table~\ref{tab6}, where we outperform all the other models by significant margins, demonstrating the advantage of our model in dense pixel-level prediction tasks. Notably, the model with handcrafted spherical wavelets outperforms other standard models, suggesting the importance of information recovery in pooling and unpooling. Please see Appendix C for more ablation study. \vspace{-0.15cm}

 \begin{table}[!t]
    \centering
    \caption{Comparison of different pooling for spherical image classification on different datasets.}\vspace{-0.15cm}
    \setlength\tabcolsep{3pt}
    \resizebox{1.0\columnwidth}{!}{\begin{tabular}{c|ccccc|ccccc}
        \hline\hline
          &\multicolumn{5}{|c|}{SMNIST}&\multicolumn{5}{|c}{SModelNet40}\\
          \hline
          Models&DownSamp&Mean&Max&LiftHS-CNN$_{H}$&LiftHS-CNN&DownSamp&Mean&Max&LiftHS-CNN$_{H}$&LiftHS-CNN\\
         Acc (\%)& 99.28&99.55&99.54&99.50&\bf{99.60}&89.26&90.76&91.05&90.96&\bf{91.37}\\
         \hline\hline
    \end{tabular}}\vspace{-0.2cm}
    \label{tab5}
\end{table}

 \begin{table}[!t]
    \centering
    \setlength\tabcolsep{3pt}
    \caption{Comparison of different pooling/unpooling for spherical image segmentation on 2D3DS.}\vspace{-0.15cm}
    \resizebox{0.55\columnwidth}{!}{\begin{tabular}{c|ccccc}
        \hline\hline
          Models&DownSamp&Mean&Max&LiftHS-CNN$_{H}$&LiftHS-CNN\\
         mIou&0.3789&0.4110&0.4042&0.4217&\bf{0.4534}\\
         mAcc (\%)&53.85&56.82&56.05&58.84&\bf{61.87}\\
         \hline\hline
    \end{tabular}}\vspace{-0.4cm}
    \label{tab6}
\end{table}

\section{Conclusion}\vspace{-0.2cm}

We proposed LiftHS-CNN to improve hierarchical feature learning in spherical domains. Lifting-based adaptive spherical wavelets were learned for pooling and unpooling to alleviate the information loss problem of existing models, where we adopted graph attention to learn the lifting operators. The \textit{spatial locality} and \textit{vanishing moments} of resulting wavelets were also studied. Experiments on various tasks validated the superiority of LiftHS-CNN. Our future study will include the optimal graph construction, and the applications to tasks such as spherical image denoising and super-resolution.

\bibliographystyle{plain}
\bibliography{references}
\end{document}